\pdfoutput=1

\documentclass[11pt]{article}

\usepackage{acl}
\usepackage{graphicx} 
\usepackage{times}
\usepackage{latexsym,multirow}

\usepackage{microtype}
\usepackage[T1]{fontenc}

\usepackage[utf8]{inputenc}

\title{Neural Language Taskonomy: Which NLP Tasks are the most Predictive of fMRI Brain Activity?}

\author{Subba Reddy Oota$^{1,2}$, Jashn Arora$^2$, Veeral Agarwal$^2$, Mounika Marreddy$^2$ \\
\textbf{Manish Gupta$^{2,3}$ and Bapi Raju Surampudi$^2$}\\
$^1$INRIA, Bordeaux, France; $^2$IIIT Hyderabad, India; $^3$Microsoft, India\\
\texttt{\small subba-reddy.oota@inria.fr, jashn.arora@research.iiit.ac.in}\\
\texttt{\small veeral.agarwal@research.iiit.ac.in, mounika.marreddy@research.iiit.ac.in}\\
\texttt{\small gmanish@microsoft.com, raju.bapi@iiit.ac.in}
}

\begin{document}
\maketitle
\begin{abstract}

Several popular Transformer based language models have been found to be successful for text-driven brain encoding.  
However, existing literature leverages only pretrained text Transformer models and has not explored the efficacy of task-specific learned Transformer representations.
In this work, we explore transfer learning from representations learned for ten popular natural language processing tasks (two syntactic and eight semantic) for predicting brain responses from two diverse datasets: Pereira (subjects reading sentences from paragraphs) and Narratives (subjects listening to the spoken stories).
Encoding models based on task features are used to predict activity in different regions across the whole brain.
Features from coreference resolution, NER, and shallow syntax parsing explain greater variance for the reading activity. 
On the other hand, for the listening activity, tasks such as paraphrase generation, summarization, and natural language inference show better encoding performance.
Experiments across all 10 task representations provide the following cognitive insights: (i) language left hemisphere has higher predictive brain activity versus language right hemisphere, (ii) posterior medial cortex, temporo-parieto-occipital junction, dorsal frontal lobe have higher correlation versus early auditory and auditory association cortex, (iii) syntactic and semantic tasks display a good predictive performance across brain regions for reading and listening stimuli resp.
\end{abstract}

\section{Introduction}
Brain encoding aims at constructing neural brain activity given an input stimulus. 
Since the discovery of the relationship between language stimuli and functions of brain networks using fMRI~[for ex., \citep{constable2004sentence}], researchers have been interested in understanding how the neural encoding models predict the fMRI brain activity. Several brain encoding models have been developed to (i) understand the ventral stream in biological vision~\citep{yamins2014performance,kietzmann2019recurrence,bao2020map}, and (ii) to study the higher-level cognition like language processing~\citep{gauthier2019linking,schrimpf2020neural,schwartz2019inducing}.

 
Some recent studies~\cite{nishida2015word,huth2016natural} have been able to identify brain ROIs (Region of Interest) that respond to words that have a similar meaning and have thus built a ``semantic atlas'' of how the human brain organizes language. Further, several studies~\cite{oota2018fmri,jain2018incorporating,hollenstein2019cognival} have used a wide variety of word embeddings where words represented as vectors in an embedding space are mapped to brain activation for improved neural coding.

Recently, Transformer~\cite{vaswani2017attention} based models like BERT~\cite{devlin2019bert} have been found to be very effective across a large number of natural language processing (NLP) tasks. These Transformer based models have been pretrained on millions of text instances in an unsupervised manner and further finetuned to specialize for various NLP tasks. Natural language understanding requires integrating several cognitive skills like syntactic parsing of the language structure, identifying the named entities, capturing the word meaning in the context, coreference resolution, etc. Learning from massive corpora enables these models to excel at cognitive skills required for language understanding. 
Interestingly, such Transformer-based neural representations have been found to be very effective for brain encoding as well~\cite{schrimpf2020neural}. 

Despite the recent advances in mapping between language Transformers and the brain activity recorded with reading~\cite{schrimpf2020neural}, the Transformer features themselves are notoriously difficult to interpret. In recent works,~\citet{caucheteux2021disentangling,antonello2021low} address this issue by disentangling the high-dimensional Transformer representations of language models into four combinatorial classes: lexical, compositional, syntactic, and semantic representations to explore which class is highly associated with language cortical ROIs. Representations do not exist in a vacuum but become meaningful only when they accomplish a task. Therefore, the next logical step is to see which of these Transformer representations most effectively drive the linear mapping between language models and the brain in the context of NLP tasks.~\citet{gauthier2019linking} 
fine-tune a pretrained BERT model on multiple tasks to find tasks best correlated with high \emph{decoding} performance. In this study, we investigate the correlation between brain activation and feature representations learned by different task-specific networks, and ask which tasks lead to
improvements in \emph{brain-encoding} performance.

Recently, a study using multiple computer vision tasks has shown that 3D vision task models predict better fMRI brain activity than 2D vision task models~\cite{wang2019neural} for visual stimuli.
Inspired by the success of correlations in the vision field~\cite{wang2019neural}, and brain encoding study of a variety of language Transformer models~\cite{schrimpf2020neural,caucheteux2021model,caucheteux2021disentangling}, we build neural language taskonomy models for brain encoding and aim to find NLP tasks that are most explanatory of brain activations for reading and listening tasks. 

In this paper, we uncover insights about the association between fMRI voxel activations and representations of diverse NLP tasks representations. 
The predictive power of task-specific representations with brain activation is ascertained by (1) using ridge regression on such representations and predicting activations and (2) computing popular metrics like 2V2 accuracy and Pearson correlation between actual and predicted activations.

Specifically, we make the following contributions in this paper. 
\begin{itemize}
    \item Given Transformer models finetuned for various NLP tasks, we propose the problem of finding which of these are the most predictive of fMRI brain activity for reading and listening tasks. 
\item  Our language taskonomy results reveal that Coreference Resolution, Named Entity Recognition, and Shallow Syntax Parsing tasks have higher predictive performance while reading the text. On the other hand, paraphrase detection, summarization, and Natural Language Inference tasks display better correlation during listening.
\item  We also perform similarity analysis between task representations from transfer learning and neural taskonomy and derive interesting cognitive insights from brain maps.
\end{itemize}

\section{Related Work}
Older methods for text-based stimulus representation include text corpus co-occurrence counts~\cite{mitchell2008predicting,pereira2013using,huth2016natural}, syntactic and discourse features~\cite{wehbe2014}. In recent times, both semantic and experiential attribute models have been explored for text-based stimuli. Semantic representation models include distributed word embeddings~\cite{pereira2016decoding,anderson2017visually,pereira2018toward,toneva2019interpreting,hollenstein2019cognival,wang2020fine}, sentence representation models~\cite{sun2019towards,toneva2019interpreting,sun2020neural}, recurrent neural networks~\cite{jain2018incorporating,oota2019stepencog}, and Transformer-based language models~\cite{gauthier2019linking,toneva2019interpreting,schwartz2019inducing,oota2022cross,oota2022visio}. 
Experiential attribute models represent words in terms of human ratings of their degree of association with different attributes of experience, typically on a scale of 0-6~\cite{anderson2019integrated,anderson2020decoding,berezutskaya2020cortical,jat2020relating,caucheteux2021disentangling,antonello2021low} or binary~\cite{handjaras2016concepts,wang2017predicting}.
Fine-grained details such as lexical, compositional, syntactic, and semantic representations of narratives are factorized from Transformer-based models and utilized for training encoding models. The resulting models are better able to disentangle the corresponding brain responses in fMRI~\cite{caucheteux2021disentangling}. 

In this paper, we focus on Transformer-based linguistic stimuli representations since they have been found to be most effective. Unlike previous studies which directly used existing task-agnostic pretrained models, we train task-specific Transformer models and aim to find which model leads to the best encoding accuracy given reading and listening language stimuli.

\section{Brain Imaging Datasets}
\label{sec:dataset}
We work with two datasets: Pereira and Narratives-Pieman. Results on Narratives-Lucy and Narratives-SlumLord show similar trends. Hence, we also show results on Narratives-Lucy and Narratives-SlumLord in the appendix.

\noindent\textbf{Pereira Dataset (Reading Sentences from Passages)} 
For the Pereira dataset, similar to earlier work~\cite{sun2019towards,sun2020neural}, we combine the data from sentence-based experiments (experiments-2 and 3) from~\citet{pereira2018toward}. Five subjects were presented a total of 627 sentences from 48 broad topics, spanning over 168 passages, where each passage consists of 3-4 sentences. 
As in~\cite{pereira2018toward}, we focused on nine brain ROIs (regions of interest) corresponding to four brain networks: (i) Default Mode Network (DMN) (linked to the functionality of semantic processing), (ii) Language Network (related to language processing, understanding, word meaning, and sentence comprehension), (iii) Task Positive Network (TP) (related to attention, salience information), and (iv) Visual Network (related to the processing of visual objects, object recognition). 
We briefly summarize the details of the dataset and the number of voxels corresponding to each ROI in Table~\ref{tab:pereira_stats}. We use the AAL parcellation Atlas (116 $\times$ 116 brain ROIs) to present the brain map results, since Pereira dataset contains annotations tied to this atlas.

\setlength{\tabcolsep}{1pt}
\begin{table}[!htb]
\scriptsize
\centering
\begin{tabular}{|l|c|c|c|c|c|c|c| c |c |}
\hline
 ROIs$\rightarrow$ & \multicolumn{2}{c|}{Language} & \multicolumn{5}{c|}{Vision} & \multicolumn{1}{c|}{DMN} & \multicolumn{1}{c|}{Task Positive} \\ \cline{1-10} 
$\downarrow$Subj& LH& RH & Body & Face& Object & Scene & Vision& RH & LH  
\\ \hline 
P01&5265&6172&3774&4963&8085&4141&12829&17190&35120 \\
M02&4930&5861&3873&4782&7552&3173&11729&15070&30594 \\
M04&5906&5401&3867&4803&7812&3602&12278&18011&34024 \\
M07&5629&5001&4190&4993&8617&3721&12454&17020&30408 \\
M15&5315&6141&4112&4941&8323&3496&12383&15995&31610 \\
\hline
\end{tabular}
\caption{\# Voxels in each ROI in the Pereira Dataset. LH - Left Hemisphere. RH - Right Hemisphere.}
\label{tab:pereira_stats}
\end{table}

\setlength{\tabcolsep}{0.5pt}
\begin{table}[!htb]
\scriptsize
\centering
\begin{tabular}{|c|c|c |c|c |c |c |c| c |c| c|}
\hline
\multirow{2}{*}{ROIs$\rightarrow$ }& \multicolumn{2}{c|}{EAC} & \multicolumn{2}{c|}{AAC} & \multicolumn{2}{c|}{PMC} & \multicolumn{2}{c|}{TPOJ}  & \multicolumn{2}{c|}{DFL}\\ 
\cline{2-11}
 & LH& RH & LH & RH& LH & RH & LH& RH & LH & RH\\ \hline 
\# Voxels& 808&638&1420&1493&1198&1204&847&1188&1061&875 \\
\hline
\end{tabular}
\caption{\# Voxels in each ROI in the Narratives Dataset. LH - Left Hemisphere. RH - Right Hemisphere. Pieman has 82, Lucy has 16 and SlumLord has 18 subjects. \# Voxels across ROIs are same for all the three.}
\label{tab:narrative_stats}
\end{table}

\noindent\textbf{Narratives-Pieman (Listening to Stories)} 
The ``Narratives'' collection aggregates a variety of fMRI datasets collected while human subjects listened to naturalistic spoken stories. The Narratives dataset that includes 345 subjects, 891 functional scans, and 27 diverse stories of varying duration totaling $\sim$4.6 hours of unique stimuli ($\sim$43,000 words) was proposed  in~\cite{nastase2021narratives}. Similar to earlier works~\cite{caucheteux2021model}, we analyze data from 82 subjects listening to the story titled `PieMan' with 259 TRs (repetition time -- fMRI recorded every 1.5 sec.). We list number of voxels per ROI in this dataset in Table~\ref{tab:narrative_stats}.
We use the multi-modal parcellation of the human cerebral cortex (Glassar Atlas: consists of 180 ROIs in each hemisphere) to display the brain maps~\cite{glasser2016multi}, since Narratives dataset contains annotations tied to this atlas.
The data covers ten brain ROIs in the human brain, i.e., Left hemisphere (L), and Right hemisphere (R) for each of the following: (i) early auditory cortex (EAC: A1, LBelt, MBelt, PBelt, and R1) which plays a key role for sound perception since it represents one of the first cortical processing stations for sounds; (ii) auditory association cortex (AAC: A4, A5, STSdp, STSda, STSvp, STSva, STGa, and TA2) which is concerned with the memory and classification of sounds; (iii) posterior medial cortex (PMC: POS1, POS2, v23ab, d23ab, 31pv, 31pd, 7m); (iv) the temporo parieto occipital junction (TPOJ: TPOJ1, TPOJ2, TPOJ3, STV, PSL) which is a complex brain territory heavily involved in several high-level neurological functions, such as language, visuo-spatial recognition, writing, reading, symbol processing, calculation, self-processing, working memory, musical memory, and face and object recognition; and (v) the dorsal frontal lobe (DFL: L\_55b, SFL, L\_44, L\_45, IFJA, IFSP) which covers the aspects of pragmatic processing such as discourse management, integration of prosody, interpretation of nonliteral meanings, inference making, ambiguity resolution, and error repair.


\section{Encoding Model}
To explore how and where contextual language features are represented in the brain when reading sentences and listening to stories, we extract different features spaces describing each stimulus sentence and use them in an encoding model to predict brain responses.
Our reasoning is as follows. If a feature is a good predictor of a specific brain region, information about that feature is likely encoded in that region.
In this paper, for both datasets, we train fMRI encoding models using Ridge regression on stimuli representations obtained using a variety of NLP tasks. The main goal of each fMRI encoder model is to predict brain responses associated with each brain region given a stimuli. In all cases, we train a model per subject separately. Following literature on brain encoding~\cite{caucheteux2021model, toneva2020modeling}, we choose to use a ridge regression model instead of more complicated models. We plan to explore more such models as part of future work. 
We follow K-fold (K=10) cross-validation. All the data samples from K-1 folds were used for training, and the model was tested on samples of the left-out fold. We used sklearn's ridge-regression with default parameters, 10-fold cross-validation, Stochastic-Average-Gradient Descent Optimizer, Huggingface for Transformer models, MSE loss function, and L2-decay ($\lambda$) as 1.0. We used BERT Word-Piece tokenizer for the linguistic Transformer input. All experiments were conducted on a machine with 1 NVIDIA GEFORCE-GTX GPU with 16GB GPU RAM. We make the code publicly available\footnote{\url{https://tinyurl.com/langTask}}.

\begin{figure*}[t] 
\centering
\includegraphics[width=\linewidth]{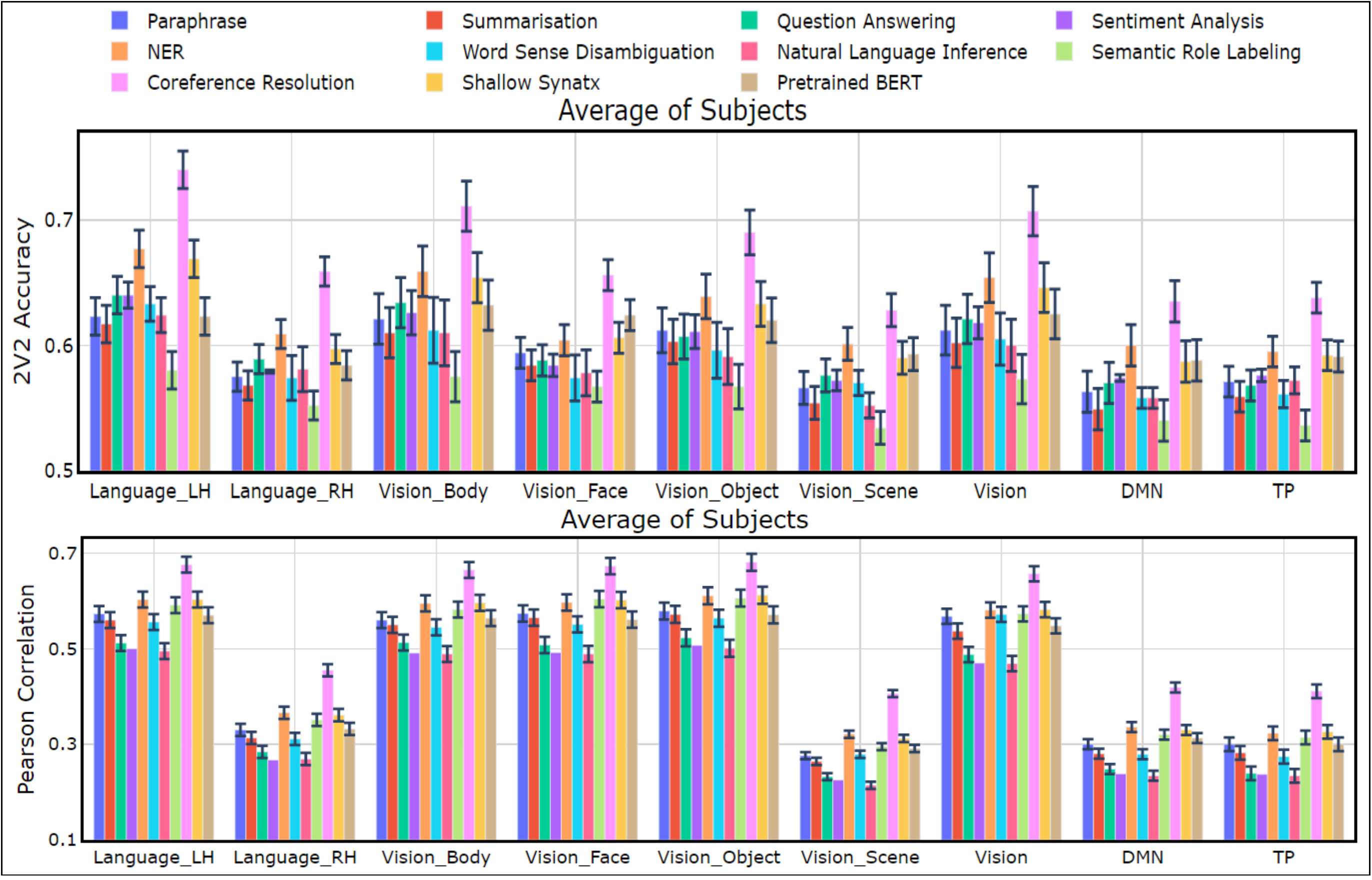}
\caption{Pereira -- 2V2 Accuracy (top figure) and Pearson correlation coefficient (bottom figure) between predicted and true responses across different brain regions using a variety of NLP tasks. Results are averaged across all participants. CR, NER, and SS perform the best.}
\label{fig:pereira_2v2_pcc_avg}
\end{figure*}

\subsection{Feature Spaces}
To simultaneously test representations from multiple NLP tasks, we used the latent space features from each of the following ten popular NLP tasks: coreference resolution (CR), named entity recognition (NER), natural language inference (NLI), paraphrase detection (PD), question answering (QA), sentiment analysis (SA), semantic role labeling (SRL), shallow syntax parsing (SS), summarization (Sum) and word sense disambiguation (WSD). All of these are discriminative NLP tasks, and thus we use models obtained by task-specific finetuning of the same pretrained Transformer encoder model (BERT-base-cased with dimensionality=768). Given an input sentence, each task Transformer outputs token representations at the final layer. We use the \#tokens $\times$ 768 dimension vector obtained from the last hidden layer to obtain latent features for the stimuli. We then build individual ridge regression models with the extracted latent features to predict brain responses and measure the correlation between the prediction and the true response.

\noindent\textbf{Pereira:} Since individual sentences were presented to the subjects while modeling, sentences were passed one by one to the task Transformer model, and average-pooled representations were used to encode the sentence stimuli.

\noindent\textbf{Narratives-Pieman:} Due to the constraint on input sequence length for BERT (512), we considered a window size of 10 sentences with the last two sentences of one window overlapping with the next to be given as input to the BERT model. We use the average-pooled representation from BERT to encode text stimuli. To get the representation for a TR, we pooled the representations of only those words of the sentences in that TR.

\subsection{Task Descriptions}
Here we describe the functionality of each NLP task that we used for fMRI encoding. \noindent\textbf{CR:} involves finding all expressions that refer to the same entity in a text. 
\noindent\textbf{PD:} involves taking a passage -- either spoken or written -- and rewording it in shorter or own words.
\noindent\textbf{Summarization (Sum):} involves selecting a few important sentences from a document or paragraph. 
\noindent\textbf{NER:} involves detection of the named entities such as person names, location names, company names from a given text.
\noindent\textbf{NLI:} investigates the entailment relationship between two texts: premise and hypothesis.
\noindent\textbf{QA:} aims to select an answer given a passage, a question, and a set of candidate answers.
\noindent\textbf{SA:} involves determining whether a piece of text is positive, negative, or neutral.
\noindent\textbf{SRL:} assigns labels to words or phrases in a sentence that indicates their semantic role in the sentence, such as that of an agent, goal, or result. 
\noindent\textbf{SS:} provides an approximation of phrase-syntactic structure of sentences.
\noindent\textbf{WSD:} involves determining which sense (meaning) of a word is activated by the use of the word in a particular context.

Syntactic reasoning is rather shallow compared to deep semantic reasoning. Syntactic reasoning follows somewhat objective grammar rules. Comparatively semantic reasoning is often subjective in nature and complex. The emerging evidence from fMRI studies~\cite{fedorenko2020lack,fedorenko2012lexical} also points out that processing of both syntax and semantics is distributed in the brain and it is only when violations of these processes are probed, we see localization of function~\cite{friederici2003role}. Thus, in this work, we explore syntactic and semantic tasks separately. Of the above mentioned tasks, NER and SS are syntactic, while the others involve semantic reasoning. 

Our selection of these tasks was based on the following design principles: (1) We wanted to select a set of tasks covering diverse cognitive-linguistic skills. (2) We wanted to select tasks that are a part of popular NLP benchmarks like GLUE~\cite{wang2018glue}. (3) We selected tasks for which BERT-base-cased finetuned models were available. Note that we did not finetune any of these models ourselves but leveraged the state-of-the-art finetuned models available on Huggingface. Details of the specific finetuned model checkpoints are mentioned in Table~\ref{tab:finetunedModelDetails} in the Appendix.

\subsection{Evaluation Metrics}
\label{sec:metrics}
We evaluate our models using popular brain encoding evaluation metrics described in the following. Given a subject and a brain region, let $N$ be the number of samples. Let $\{Y_i\}_{i=1}^N$ and $\{\hat{Y}_i\}_{i=1}^N$ denote the actual and predicted voxel value vectors for the $i^{th}$ sample. Thus, $Y\in R^{N\times V}$ and $\hat{Y}\in R^{N\times V}$ where $V$ is the number of voxels in that region.

\noindent\textbf{2V2 Accuracy} is computed as 
    2V2Acc=$\frac{1}{N_{C_2}}\sum_{i=1}^{N-1}\sum_{j=i+1}^N I[cosD(Y_i, \hat{Y}_i)+cosD(Y_j, \hat{Y}_j)<cosD(Y_i, \hat{Y}_j)+cosD(Y_j, \hat{Y}_i)]$
\noindent where $cosD$ is the cosine distance function. $I[c]$ is an indicator function such that $I[c]=1$ if $c$ is true, else it is 0. The higher the 2V2 accuracy, the better.

\noindent\textbf{Pearson Correlation (PC)} is computed as PC=$\frac{1}{N}\sum_{i=1}^{n} corr[Y_i, \hat{Y}_i]$ where corr is the correlation function.

\noindent\textbf{Mean Absolute Error (MAE)} is computed as MAE=$\frac{1}{N}\sum_{i=1}^{n} |[Y_i-\hat{Y}_i]|$.


\noindent\textbf{Statistical Significance}: In order to estimate the statistical significance of the performance differences (across all results), we performed one-way ANOVA on the mean values for the subjects. In all such cases we report p-values corrected using Bonferroni correction.

\subsection{Neural Language Tasks Similarity Computation}
To estimate the similarity between 10 language tasks, we took the prediction performance scores across all the voxels in Pereira (97,539) and Narratives-Pieman datasets (10,732). To analyze the relationship between tasks based on neural representations, we calculated the Pearson correlation between predicted voxels of each task with the remaining tasks. These Pearson correlation values were used to construct heatmaps and the task similarity trees(dendograms) using hierarchical clustering for Pereira and Narratives-Pieman datasets.

\begin{figure*}[!htb] 
\centering
\includegraphics[width=\linewidth]{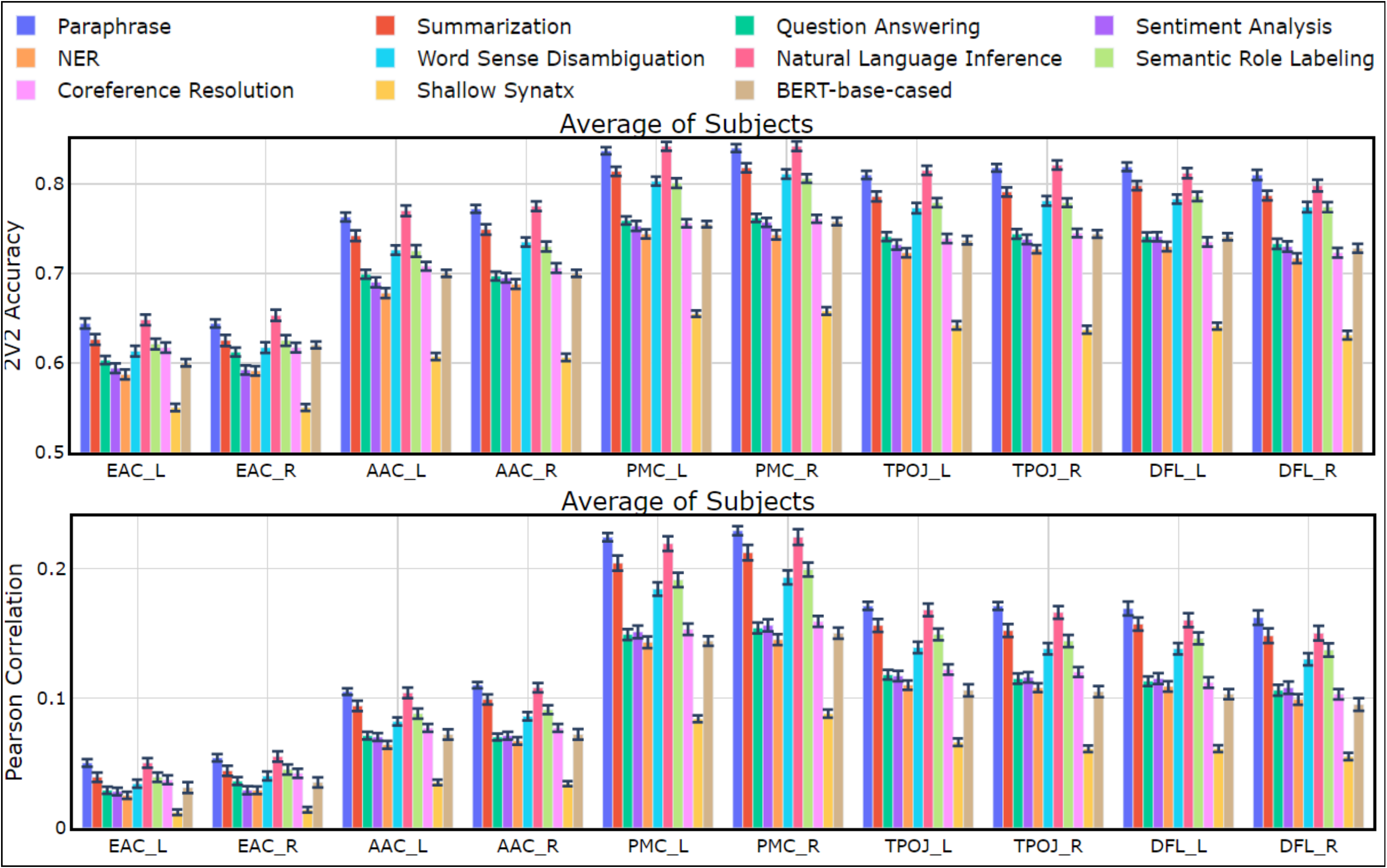}
\caption{Narratives-Pieman -- 2V2 Accuracy (top figure) and Pearson correlation coefficient (bottom figure) between predicted and true responses across different brain regions using a variety of NLP tasks. Results are averaged across all participants. NLI, PD, and Summarization perform the best.}
\label{fig:pieman_2v2_pcc_avg}
\end{figure*}

\section{Results}
In order to assess the performance of the fMRI encoder models learned using the representations from a variety of NLP tasks, we computed the 2V2 accuracy and Pearson correlation coefficient between the predicted and true responses across various ROIs for both the reading (Pereira) dataset (Fig.~\ref{fig:pereira_2v2_pcc_avg}) as well as the listening (Narratives-Pieman) dataset (Fig.~\ref{fig:pieman_2v2_pcc_avg}). 

\subsection{Encoding performance of Language Task models for reading vs listening tasks}
\noindent\textbf{Reading Sentences (Pereira):} From Fig.~\ref{fig:pereira_2v2_pcc_avg}, we observe that tasks such as CR, NER, SRL, and SS appear to have a better correlation to the brain responses compared to the other tasks. In order to estimate the statistical significance of the performance differences, we performed one-way ANOVA on the mean correlation values for the subjects across the ten language tasks for the nine brain ROIs. The main effect of the ANOVA test was significant for all the ROIs with p$\leq 10^{-2}$ with confidence 95\% (see Appendix for detailed ANOVA results). Further, \emph{post hoc} pairwise comparisons~\cite{ruxton2008time} confirmed the visual observations that on both 2V2 accuracy and Pearson correlation measures, tasks such as CR, NER, SRL, and SS performed significantly better compared to other tasks (see Appendix for pairwise comparison results). These results demonstrate that when reading a sentence, information processing operations related to recognizing named entities, labeling semantic roles to the constituents of a sentence, identifying the references from a sentence to the given topic (concept), and syntactic processing may be engaged. 

Further, we observe that the ROI corresponding to language processing in the left hemisphere (Language\_LH) has higher encoding performance than that of the right hemisphere (Language\_RH). This is in line with the left hemisphere dominance for language processing~\cite{binder2009semantic}. Also, lateral visual ROIs such as Vision\_Object, Vision\_Body, Vision\_Face, and Vision ROIs display higher correlation with the language tasks associated with named entities (NER), relating the entities (CR), and syntax processing (SS). Higher correlations with all the visual brain regions point to the possible alignment of visual and language regions for semantic understanding~\cite{popham2021visual} in a reading task. Finally, across all regions, pretrained BERT model has worse correlation compared to at least 5 other task models.

\noindent\textbf{Listening Stories (Narratives-Pieman):} From Fig.~\ref{fig:pieman_2v2_pcc_avg}, we observe that the profiles of performance show low scores in the early auditory cortex (EAC), auditory association cortex (AAC); average scores in TPOJ and DFL; and superior scores in PMC. This aligns with the known language hierarchy for spoken language understanding~\cite{nastase2020leveraging}. Tasks such as PD, Summarization, and NLI seem to yield better performance in predicting the brain responses than the other NLP tasks across all the ROIs. 
These Pearson correlation ($\tau$)  results are comparatively much higher compared to those obtained using pretrained (task-agnostic) GPT2 model in~\cite{caucheteux2021disentangling} ($\tau$ ranging from $0.02-0.06$).
As shown in Fig.~\ref{fig:pieman_2v2_pcc_avg}, our method obtains much higher correlations ($\tau$ ranging from $0.02-0.229$). Similar to the Pereira dataset, we estimate the statistical significance of the performance differences using the one-way ANOVA test. The main effect of task was significant for all the ROIs with p$\leq 10^{-3}$ with confidence 95\% (see Appendix for detailed ANOVA results). Also, \emph{Post hoc} pairwise comparisons~\cite{ruxton2008time} revealed that on both 2V2 accuracy and Pearson correlation measures, tasks such as PD, Sum, and NLI performed significantly better compared to other tasks (see Appendix for pairwise comparison results).

Further, from Fig.~\ref{fig:pieman_2v2_pcc_avg}, we see that the bilateral posterior medial cortex (PMC) associated with higher language function exhibits a higher correlation among all the brain ROIs. ROIs, including bilateral TPOJ and bilateral DFL, yield higher correlations with the five NLP tasks, which is in line with the language processing hierarchy in the human brain. Finally, across all regions, pretrained BERT model has worse correlation compared to at least 5 other task models.

In summary, different and distinct language Taskonomy features seem to be related to the encoding performance in reading versus listening tasks. CR, NER, SRL, and SS perform better for reading. PD, Sum, and NLI perform better for listening. While listening the subject is cognitively more involved in the activity compared to reading~\cite{buchweitz2009brain}. Thus, it makes sense that shallow tasks like NER and SS are useful for reading while more complex NLP tasks like PD, Sum and NLI are effective for encoding listening stimuli.

\begin{figure}[t] 
\centering
\includegraphics[width=\linewidth]{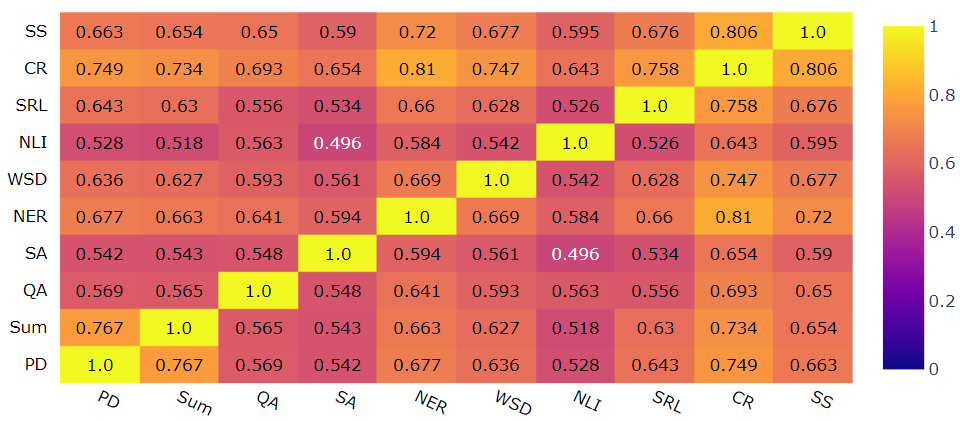}
\vspace{-10pt}
\caption{Pereira -- Prediction Similarity Matrix constructed from the task-wise brain response predictions across 10 tasks averaged across all subjects.}
\label{fig:tasksimilarity_pereira}
\end{figure}

\begin{figure}[!htb] 
\centering
\includegraphics[width=\linewidth]{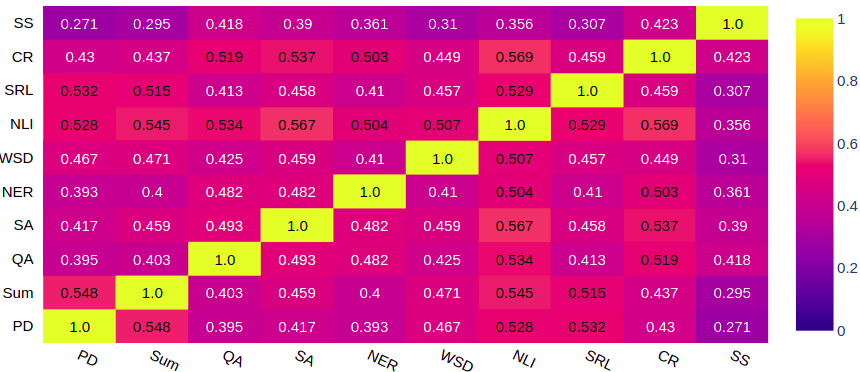}
\vspace{-10pt}
\caption{Narratives-Pieman -- Prediction Task Similarity constructed from the task-wise brain response predictions across 10 tasks averaged across all subjects.}
\label{fig:tasksimilarity_narratives}
\end{figure}




\begin{figure}[t]
\includegraphics[width=0.49\linewidth]{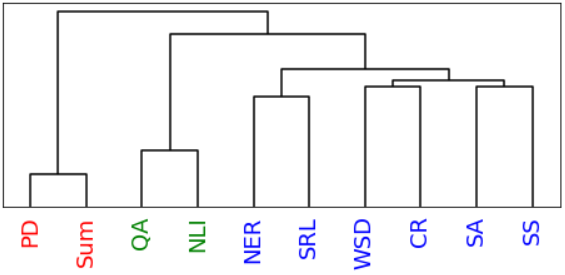}
\includegraphics[width=0.49\linewidth]{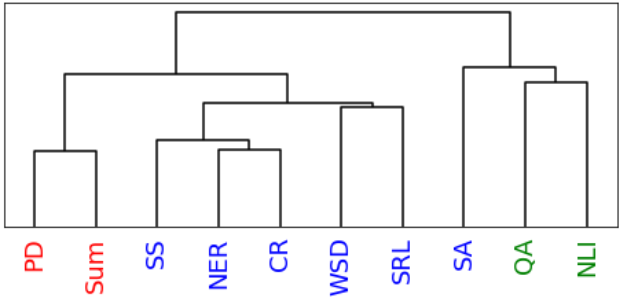}
\caption{Left: Pereira Dendrogram constructed using similarity on representations from task-specific Transformer encoder models with stimuli from the dataset passed as input. Right: Pereira Dendrogram constructed using similarity matrix shown in Fig.~\ref{fig:tasksimilarity_pereira}.}
\label{fig:Dendrograms_pereira_transfer}
\end{figure}

\begin{figure}[t]
\includegraphics[width=0.49\linewidth]{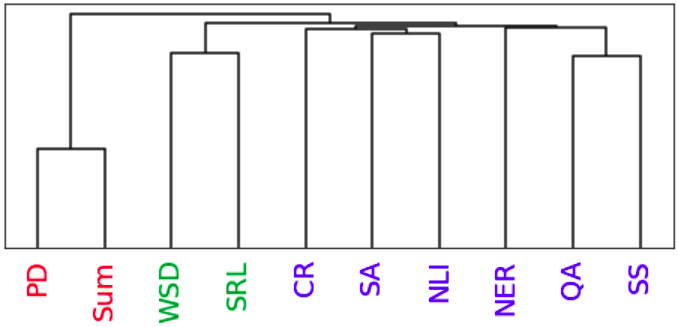}
\includegraphics[width=0.49\linewidth]{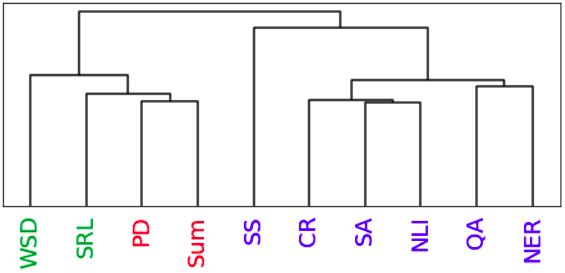}
\caption{Left: Narratives-Pieman Dendrogram constructed using similarity on representations from task-specific Transformer encoder models with stimuli from the dataset passed as input. Right: Narratives-Pieman Dendrogram constructed using similarity matrix shown in Fig.~\ref{fig:tasksimilarity_narratives}.}
\label{fig:Dendrograms_narratives_transfer}
\end{figure}

\begin{figure*}[t]
\centering
\includegraphics[width=\linewidth]{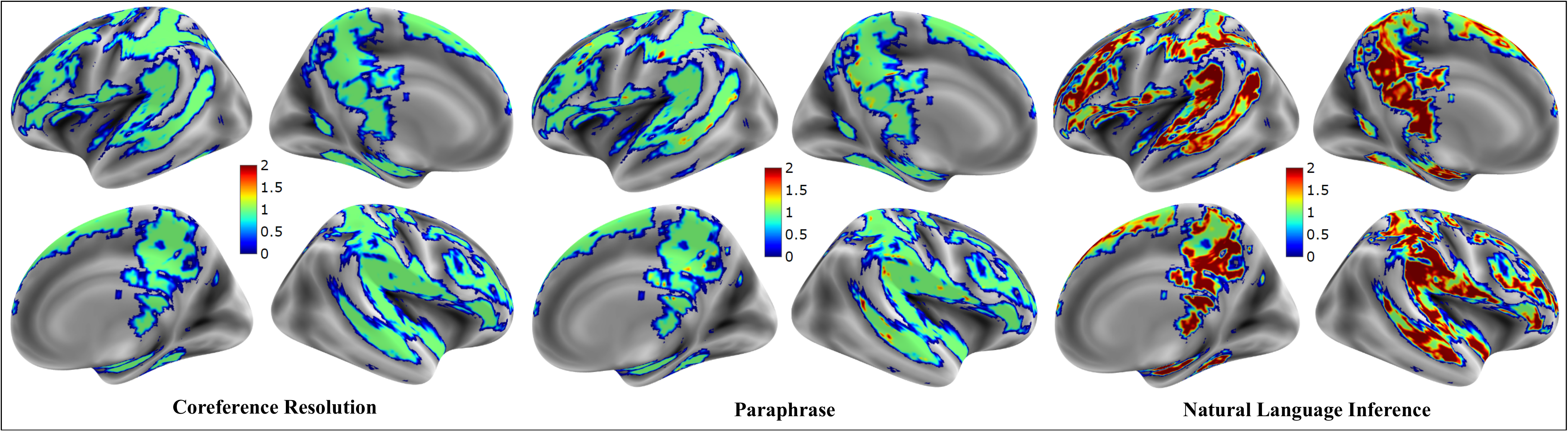}
\caption{Pereira BrainMaps: Mean absolute error (MAE) between predictive voxels and actual voxels using task features from Taskonomy in one sample subject (subject 1). Predictive regions of different tasks are dissimilar across tasks. The MAE values of each brain ROI are: CR  (Language: 0.64, Visual: 0.57, DMN: 1.19, TP: 0.67), PD (Language: 0.81, Visual: 0.74, DMN: 1.34, TP: 0.87) and NLI (Language: 1.9, Visual: 1.88, DMN: 2.1, TP: 2.03).}
\label{fig:brainmaps_pereira}
\end{figure*} 

\begin{figure*}[t] 
\centering
\includegraphics[width=\linewidth]{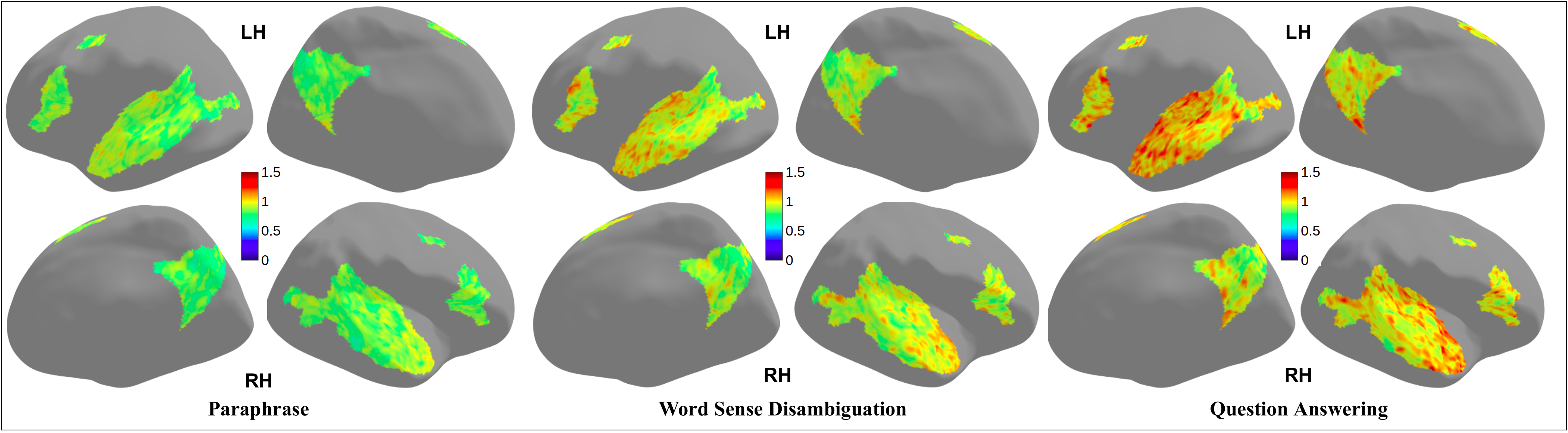}
\caption{Narratives-Pieman BrainMaps: Mean absolute error (MAE) between predictive voxels and actual voxels using task features from Taskonomy in one sample subject (subject 1) of PieMan dataset. Predictive regions of various tasks are different across tasks. The MAE values of each brain ROI: PD task (EAC: 0.74, AAC: 0.66, PMC: 0.60, TPOJ: 0.61, and DFL: 0.694), WSD task(EAC: 0.83, AAC: 0.75, PMC: 0.68, TPOJ: 0.68, and DFL: 0.76), QA task (EAC: 0.92, AAC: 0.83, PMC: 0.74, TPOJ: 0.75, and DFL: 0.76).}
\label{fig:brainmaps_narratives}
\end{figure*}

\subsection{Language Task Similarity Computation}
Pearson correlation values between predicted responses for each pair of tasks were used to construct the similarity matrix with heatmap for both Pereira and Narratives-Pieman datasets, as shown in Figs.~\ref{fig:tasksimilarity_pereira} and~\ref{fig:tasksimilarity_narratives}. We observe that the following task pairs are highly correlated for the Pereira dataset: (NER and CR), (SS and CR) and (PD and Sum). Also these task pairs are highly correlated for the Narratives-Pieman dataset: (CR and NLI), (NLI and SA) and (PD and Sum). Similarities are relatively higher for Narratives-Pieman compared to the Pereira dataset. Surprisingly, the (NLI, SA) pair has lowest similarity for Pereira (reading) and close to highest in Narratives-Pieman (listening). We hypothesize that this is because sentiment is best conveyed while the subject is listening.
 
\noindent\textbf{Reading sentences (Pereira):}
The stimulus sentences from the Pereira dataset were fed as input to each of the 10 task Transformers. The similarity among the resulting representations was analyzed using hierarchical clustering, and the clusters are visualized as dendrograms in Fig.~\ref{fig:Dendrograms_pereira_transfer} (left). We observe that the tasks are clustered into three groups denoted using red, green, and blue colors. Next, we wished to check if similar task grouping is observed on brain activations predicted by ridge regression trained on task-specific representations. Hence, similar clustering analysis was conducted on the neural space representations, and the clusters are visualized as dendrograms in Fig.~\ref{fig:Dendrograms_pereira_transfer} (right) across all subjects. Interestingly, the tree derived from brain representation also shows a similar distribution of tasks across the three groups. Similar dendrograms for individual subjects are illustrated in Appendix-Fig.~\ref{fig:Dendrograms_pereira_transfer_subjects}.

\noindent\textbf{Listening Stories (Narratives-Pieman):}
Fig.~\ref{fig:Dendrograms_narratives_transfer} compares the task similarity tree based on the patterns from the pretrained task Transformers, with the task similarity tree generated based on similarity in brain response prediction performance averaged across all subjects. We observe that the tasks are clustered into three groups denoted using red, green, and blue colors. Again, the tree derived from brain representation also shows a similar distribution of tasks across the three groups. Dendrograms for individual subjects are in the Appendix-Fig.~\ref{fig:Dendrograms_narratives_transfer_subjects}.

\subsection{Brain maps for whole brain predictions}
The mean absolute error (MAE) between predictive and actual responses is obtained using individual task features from the taskonomy. MAE values are obtained for all the voxels in the brain for both the reading (Fig.~\ref{fig:brainmaps_pereira}) and listening datasets (Fig.~\ref{fig:brainmaps_narratives}).

In the {\bf reading task}, we observe from Fig.~\ref{fig:brainmaps_pereira} that CR has lower MAE compared to PD which in turn has lower MAE compared to the  NLI task (brain maps for the other tasks are reported in Fig.~\ref{fig:brainmaps_pereira_all} in the Appendix). Overall, for the reading stimuli, tasks such as NLI, QA, and SA display higher MAE values. To further investigate which sub ROIs (LPTG, LMTG, LATG, LFus, Lpar, Lang, LIFGorb, LIFG, LaMFG, LpMFG, and LmMFG) of the Language network 
are related to the predictive task features, we train encoding models for all the sub ROIs for the best encoding task, i.e., for the CR task (see Fig.~\ref{fig:language_pereira} in Appendix). We notice that both LMTG (middle temporal gyrus) and LPTG (posterior temporal gyrus) are more accurately predicted than the other sub ROIs. On the other hand, LIFG-orb displays a lower Pearson correlation for the CR task. The presence of superior encoding information in the ROIs in the temporal gyrus as compared to those in the inferior frontal gyrus seems to mirror similar observations seen in decoder performance~\cite{anderson2017predicting}.

On the other hand, in the {\bf listening task}, we observe from Fig.~\ref{fig:brainmaps_narratives} that Paraphrase and WSD display lower MAE values compared to QA task (brain maps for the other tasks are reported in Fig.~\ref{fig:brainmaps_narratives_all} in the Appendix). Taken together, for listening stimuli, tasks such as NER, QA, SA, CR, and SS display higher MAE values.
From Fig.~\ref{fig:brainmaps_narratives}, we see that ROIs such as EAC and AAC have higher MAE compared to PMC and TPOJ brain ROIs.

We further demonstrate the prediction performance of the encoder model trained on sub ROIs for the paraphrase task in Fig.~\ref{fig:subrois_narratives} in the Appendix. It can be observed that sub ROIs such as Pos1 and Pos2 have a higher Pearson correlation than other sub ROIs of the PMC region. Both sfl and l55b display a higher correlation among all the sub ROIs for the DFL ROI. However, all the sub ROIs in the TPOJ yield higher correlation, as shown in Fig.~\ref{fig:subrois_narratives}. The control and attention ROIs in the posterior cingulate cortex (for ex., POS1 in PMC), together with the superior frontal language region (sfl in DFL) and TPOJ, are part of the well-known language network associated with narrative comprehension~\cite{nastase2020leveraging}, and it is heartening to see that task features from PD task also relate to semantic analysis of the ongoing narrative.

\subsection{Discussion}
(1) We used a ridge regression model instead of more complicated models for encoding. We believe that more complex models can lead to further exciting insights. (2) We experimented with 10 NLP tasks. Models can be pretrained for more such tasks to check if other tasks are better predictive of voxel activations. (3) We leveraged models finetuned using datasets of different sizes across tasks. While a fair comparison of dataset sizes across tasks is impossible, we understand that this could have resulted in some bias in our results. (4) We used a different dataset for reading vs listening. While we believe that the differences in task-specific model performances across reading and listening are mainly due to the learned stimulus representations, but they could also arise from other factors such as experimental conditions, the text domain of the stimuli or number of voxels, etc. (5) On Natural Language Understanding tasks such as NLI, SA, QA and PD,~\citet{gauthier2019linking} observed that scrambled sentence representations gave better decoding performance. But encoding models (especially for the listening task), scrambled order would be detrimental to making sense of what is being heard. It is an interesting future task to see if the opposite result is seen in the case of brain encoding models. It is plausible that brain uses encoding models in a flexible way when it comes to decoding~\cite{kriegeskorte2019interpreting}.~\citet{kriegeskorte2019interpreting} mention that ``Decoding models can help reveal whether particular information is present in a brain region in a format the decoder can exploit. Encoding models make comprehensive predictions about representational spaces.'' In this sense, results of current work are not directly comparable to those of~\citet{gauthier2019linking}.


\section{Conclusion}
In this paper, we studied the effectiveness of task specific NLP models for brain encoding. We observe that building individual encoding models and exploiting existing relationships among models can provide a more in-depth understanding of the neural representation of language information. Our experiments on Pereira and Narrative datasets lead to interesting cognitive insights. 

\newpage
\section{Ethical Statement}
We reused publicly available datasets for this work: Pereira and Narratives. We did not collect any new dataset. 

Pereira dataset can be downloaded from \url{https://osf.io/crwz7/}. Please read their terms of use\footnote{\url{https://github.com/CenterForOpenScience/cos.io/blob/master/TERMS_OF_USE.md}} for more details.

Narratives dataset can be dowloaded from \url{https://datasets.datalad.org/?dir=/labs/hasson/narratives}. Please read their terms of use\footnote{\url{https://datasets.datalad.org/labs/hasson/narratives/stimuli/README}} for more details.

We do not foresee any harmful uses of this technology. 

\bibliography{references}
\bibliographystyle{acl_natbib}
\appendix

\section{Details of the Finetuned Models}
We selected tasks for which BERT-base-cased finetuned models were available. Note that we did not finetune any of these models ourselves but leveraged the state-of-the-art finetuned models available on Huggingface. Details of the specific finetuned model checkpoints are mentioned in Table~\ref{tab:finetunedModelDetails}.

\begin{table*}
    \centering
    \scriptsize
    \begin{tabular}{|l|l|l|p{2in}|}
       \hline
       Task&HuggingFace Model Name & Dataset & URL\\
       \hline
              \hline
           NLI&
bert-base-nli-mean-tokens&Stanford Natural Language Inference (SNLI), MultiNLI&\url{https://huggingface.co/sentence-transformers/bert-base-nli-mean-tokens}\\
           \hline
           PD&bert-base-cased-finetuned-mrpc&Microsoft Research Paraphrase Corpus (MRPC)&\url{https://huggingface.co/bert-base-cased-finetuned-mrpc}\\
           \hline
           SS&bert-base-chunl&CoNLL-2003&\url{https://huggingface.co/vblagoje/bert-english-uncased-finetuned-chunk}\\
           \hline
           Sum&bart-base-samsum&SAMSum&\url{https://huggingface.co/lidiya/bart-base-samsum}\\
           \hline
           WSD&bert-base-baseline &English all-words &\url{https://github.com/BPYap/BERT-WSD}\\
           \hline
           CR&
bert\_coreference\_base & OntoNotes and GAP&\url{https://github.com/mandarjoshi90/coref} \\
           \hline
           NER&bert-base-NER & CoNLL-2003&\url{https://huggingface.co/dslim/bert-base-NER}\\
           \hline
           QA&bert-base-qa & SQUAD & \url{https://huggingface.co/docs/transformers/model\_doc/bert\#bertforquestionanswering}\\
           \hline
           SA&bert-base-sst&Stanford Sentiment Treebank (SST)&\url{https://huggingface.co/barissayil/bert-sentiment-analysis-sst}\\
           \hline
           SRL&bert-base-srl&English PropBank SRL&\url{https://s3-us-west-2.amazonaws.com/allennlp/models/bert-base-srl-2020.02.10.tar.gz}\\
          \hline 
    \end{tabular}
    \caption{Details of the finetuned models}
    \label{tab:finetunedModelDetails}
\end{table*}

\section{ANOVA test results}

\subsection{Pereira dataset}
The main effect of model was significant for the ROIs with 95\% confidence with these statistics:

\begin{itemize}
\itemsep0em
\item Language\_LH: [F(9, 40) = 3.95, \emph{p}=0.0052]
\item Language\_RH: [F(9, 40) = 4.53, \emph{p}=0.0015]
\item Vision\_Body: [F(9, 40) = 4.397, \emph{p}=0.00227]
\item Vision\_Face: [F(9, 40) = 3.46, \emph{p}=0.0085]
\item Vision\_Object: [F(9, 40) = 3.40, \emph{p}=0.0121]
\item Vision\_Scenes: [F(9, 40) = 4.917, \emph{p}=0.0007]
\item Vision: [F(9, 40) = 3.945, \emph{p}=0.00385]
\item DMN: [F(9, 40) = 6.28, \emph{p}=0.00034]
\item TP: [F(9, 40) = 6.54, \emph{p}=0.00042]
\end{itemize}

\begin{table}[!htb]
\scriptsize
\centering
\begin{tabular}{|c|c|c |}
\hline
T1 & T2& p-value\\ \hline 
CR & QA & 0.024 \\ \hline
CR & SA & 0.015 \\ \hline
CR & NLI & 0.010 \\
\hline
\end{tabular}
\caption{Pairwise comparison one-way ANOVA results for Language\_LH region}
\label{tab:language_lh}
\end{table}

\begin{table}[!htb]
\scriptsize
\centering
\begin{tabular}{|c|c|c |}
\hline
T1 & T2& p-value\\ \hline 
CR & SS & 0.021 \\ \hline
CR & SRL & 0.0003 \\ \hline
CR & Sum & 0.003 \\ \hline
CR & QA & 0.039 \\ \hline
CR & SA & 0.013 \\ \hline
CR & WSD & 0.016 \\
\hline
\end{tabular}
\caption{Pairwise comparison one-way ANOVA results for Language\_RH region}
\label{tab:language_rh}
\end{table}

\begin{table}[!htb]
\scriptsize
\centering
\begin{tabular}{|c|c|c |}
\hline
T1 & T2& p-value\\ \hline 
CR & SRL & 0.0011 \\ \hline
CR & Sum & 0.0092 \\ \hline
CR & SA & 0.039 \\ \hline
CR & NLI & 0.0061 \\ 
\hline
\end{tabular}
\caption{Pairwise comparison one-way ANOVA results for Vision\_body region}
\label{tab:vision_body}
\end{table}

\begin{table}[!htb]
\scriptsize
\centering
\begin{tabular}{|c|c|c |}
\hline
T1 & T2& p-value\\ \hline 
CR & SA & 0.0404 \\ \hline
CR & nli & 0.036 \\  
\hline
\end{tabular}
\caption{Pairwise comparison one-way ANOVA results for Vision\_face region}
\label{tab:vision_face}
\end{table}

\begin{table}[!htb]
\scriptsize
\centering
\begin{tabular}{|c|c|c |}
\hline
T1 & T2& p-value\\ \hline 
CR & SRL & 0.0027 \\ 
\hline
\end{tabular}
\caption{Pairwise comparison one-way ANOVA results for Vision\_object region}
\label{tab:vision_object}
\end{table}

\begin{table}[!htb]
\scriptsize
\centering
\begin{tabular}{|c|c|c |}
\hline
T1 & T2& p-value\\ \hline 
CR & Sum & 0.027 \\ \hline
CR & QA & 0.0036 \\ \hline
CR & SA & 0.0022 \\ \hline
CR & NLI & 0.0010 \\ 
\hline
\end{tabular}
\caption{Pairwise comparison one-way ANOVA results for Vision\_scene region}
\label{tab:vision_scene}
\end{table}

\begin{table}[!htb]
\scriptsize
\centering
\begin{tabular}{|c|c|c |}
\hline
T1 & T2& p-value\\ \hline 
CR & SRL & 0.0014 \\ \hline
CR & Sum & 0.0431 \\ \hline
CR & NLI & 0.0177 \\ 
\hline
\end{tabular}
\caption{Pairwise comparison one-way ANOVA results for Vision region}
\label{tab:vision}
\end{table}

\begin{table}[!htb]
\scriptsize
\centering
\begin{tabular}{|c|c|c |}
\hline
T1 & T2& p-value\\ \hline 
CR & NLI & 0.027\\ \hline
CR & Sum & 0.008\\ \hline
CR & PD & 0.0147\\ \hline

NLI & SA & 0.056\\ \hline
NLI & SS & 0.000011\\ \hline

SA & Sum & 0.0188\\ \hline
SA & PD & 0.032\\ \hline

SS & Sum & 0.000002\\ \hline
SS & WSD & 0.0059\\ \hline
SS & PD & 0.000004\\ \hline

SRL & Sum & 0.0545\\ \hline
SRL & PD & 0.08876 \\
\hline
\end{tabular}
\caption{Pairwise comparison one-way Anova results for EAC-L region}
\label{tab:EAC-L}
\end{table}

\begin{table}[!htb]
\scriptsize
\centering
\begin{tabular}{|c|c|c |}
\hline
T1 & T2& p-value\\ \hline 
NLI & SS & 0.00157\\ \hline
Sum & SS & 0.0015\\ \hline
PD & SS & 0.002\\ \hline

SA & SS & 0.0565\\ \hline
SS & WSD & 0.052\\
\hline
\end{tabular}
\caption{Pairwise comparison one-way Anova results for EAC-R region}
\label{tab:EAC-R}
\end{table}

\begin{table}[!htb]
\scriptsize
\centering
\begin{tabular}{|c|c|c |}
\hline
T1 & T2& p-value\\ \hline 
NLI & SS & 0.000007\\ \hline
SA & SS & 0.029\\ \hline
SS & SRL & 0.0084\\ \hline
SS & PD & 0.000023\\ \hline
SS & QA & 0.00128\\
\hline
\end{tabular}
\caption{Pairwise comparison one-way Anova results for AAC-L region}
\label{tab:AAC-L}
\end{table}

\begin{table}[!htb]
\scriptsize
\centering
\begin{tabular}{|c|c|c |}
\hline
T1 & T2& p-value\\ \hline 
CR & NLI & 0.0203\\ \hline
CR & PD & 0.0072\\ \hline
NER & PD & 0.0291\\ \hline
NLI & SS & 0.0000013\\ \hline
SA & SS & 0.0299\\ \hline
SS & SRL & 0.0011\\ \hline
SS & PD & 2.97929e-7\\ \hline
SS & WSD & 0.0444\\ \hline
SS & QA & 0.00099\\ \hline
PD & Sum & 0.039\\ 
\hline
\end{tabular}
\caption{Pairwise comparison one-way Anova results for AAC-R region}
\label{tab:AAC-R}
\end{table}

\begin{table}[!htb]
\scriptsize
\centering
\begin{tabular}{|c|c|c |}
\hline
T1 & T2& p-value\\ \hline 
CR & NER & 1.07034e-10\\ \hline
CR & NLI & 0.000001\\ \hline
CR & SRL & 0.0014\\ \hline
CR & PD & 0.0000047\\ \hline
CR & QA & 0.0023\\ \hline
NER & NLI & 9.02023e-11\\ \hline
NER & SA & 9.02993e-11\\ \hline
NER & SS & 0.000157159\\ \hline
NER & SRL & 9.02116e-11\\ \hline
NER & PD & 9.02023e-11\\ \hline
NER & Sum & 9.03064e-11\\ \hline
NER & WSD & 9.03172e-11\\ \hline
NER & QA & 9.02116e-11\\ \hline
NLI & SA & 0.0207013\\ \hline
NLI & SS & 9.03255e-11\\ \hline
NLI & Sum & 0.0043\\ \hline
NLI & WSD & 0.00036\\ \hline
SA & SS & 0.0000072\\ \hline
SS & SRL & 4.47012e-10\\ \hline
SS & PD & 9.04392e-11\\ \hline
SS & Sum & 0.00011\\ \hline
SS & WSD & 0.00084\\ \hline
SS & QA & 6.36666e-10\\ \hline
PD & Sum & 0.012\\ \hline
PD & WSD & 0.0012 \\
\hline
\end{tabular}
\caption{Pairwise comparison one-way Anova results for PMC-L region}
\label{tab:PMC-L}
\end{table}

\begin{table}[!htb]
\scriptsize
\centering
\begin{tabular}{|c|c|c |}
\hline
T1 & T2& p-value\\ \hline 
CR & NER & 1.52787e-9\\ \hline
CR & NLI & 0.0000042\\ \hline
CR & SS & 0.0039\\ \hline
CR & PD & 0.00011\\ \hline
CR & QA & 0.0101\\ \hline
NER & NLI & 8.86012e-11\\ \hline
NER & SA & 8.87732e-11\\ \hline
NER & SRL & 8.88714e-11\\ \hline
NER & PD & 8.86092e-11\\ \hline
NER & Sum & 1.05034e-10\\ \hline
NER & WSD & 1.01319e-10\\ \hline
NER & QA & 8.86657e-11\\ \hline
NLI & SA & 0.0059\\ \hline
NLI & SS & 8.87066e-11\\ \hline
NLI & Sum & 0.000371\\ \hline
NLI & WSD & 0.000191\\ \hline
SA & SS & 0.0000021\\ \hline
SS & SRL & 0.00000142\\ \hline
SS & PD & 8.87554e-11\\ \hline
SS & Sum & 0.000126402\\ \hline
SS & WSD & 0.000128239\\ \hline
SS & QA & 1.31249e-10\\ \hline
PD & Sum & 0.00619\\ \hline
PD & WSD & 0.0036\\
\hline
\end{tabular}
\caption{Pairwise comparison one-way Anova results for PMC-R region}
\label{tab:PMC-R}
\end{table}

\begin{table}[!htb]
\scriptsize
\centering
\begin{tabular}{|c|c|c |}
\hline
T1 & T2& p-value\\ \hline 
CR & NLI & 0.00069\\ \hline
CR & PD & 0.00395\\ \hline
NER & NLI & 0.0051\\ \hline
NER & SS & 0.0286244\\ \hline
NER & PD & 0.0235\\ \hline
NLI & SS & 4.43074e-10\\ \hline
NLI & Sum & 0.0068987\\ \hline
NLI & WSD & 0.02709\\ \hline
SA & SS & 0.0001732\\ \hline
SS & SRL & 0.0000530\\ \hline
SS & PD & 4.37008e-9\\ \hline
SS & Sum & 0.0219850\\ \hline
SS & WSD & 0.005447\\ \hline
SS & QA & 0.0000016\\ \hline
PD & Sum & 0.0306\\
\hline
\end{tabular}
\caption{Pairwise comparison one-way Anova results for TPOJ-L region}
\label{tab:TPOJ-L}
\end{table}

\begin{table}[!htb]
\scriptsize
\centering
\begin{tabular}{|c|c|c |}
\hline
T1 & T2& p-value\\ \hline 
CR & NLI & 0.0064\\ \hline
CR & PD & 0.0148564\\ \hline
NER & NLI & 0.0449\\ \hline
NLI & SS & 3.74353e-8\\ \hline
NLI & WSD & 0.0321627\\ \hline
SA & SS & 0.0036278\\ \hline
SS & SRL & 0.001054\\ \hline
SS & PD & 1.33146e-7\\ \hline
SS & Sum & 0.025420\\ \hline
SS & QA & 0.000049\\ 
\hline
\end{tabular}
\caption{Pairwise comparison one-way Anova results for TPOJ-R region}
\label{tab:TPOJ-R}
\end{table}

\begin{table}[!htb]
\scriptsize
\centering
\begin{tabular}{|c|c|c |}
\hline
T1 & T2& p-value\\ \hline 
CR & NLI & 0.000032\\ \hline
CR & PD & 0.000019\\ \hline
NER & NLI & 0.000619887\\ \hline
NER & SS & 0.040\\ \hline
NER & PD & 0.000399\\ \hline
NLI & SS & 1.61916e-10\\ \hline
NLI & Sum & 0.00074\\ \hline
NLI & WSD & 0.000462932\\ \hline
SA & SS & 0.000221241\\ \hline
SS & SRL & 0.0000123345\\ \hline
SS & PD & 1.30279e-10\\ \hline
SS & Sum & 0.0356814\\ \hline
SS & WSD & 0.0496343\\ \hline
SS & QA & 0.00000162\\ \hline
PD & Sum & 0.0004803\\ \hline
PD & WSD & 0.000296713\\ 
\hline
\end{tabular}
\caption{Pairwise comparison one-way Anova results for DFL-L region}
\label{tab:DFL-L}
\end{table}

\begin{table}[!htb]
\scriptsize
\centering
\begin{tabular}{|c|c|c |}
\hline
T1 & T2& p-value\\ \hline 
CR & NLI & 0.000191\\ \hline
CR & PD & 0.00010\\ \hline
NER & NLI & 0.0168115\\ \hline
NER & SS & 0.0168115\\ \hline
NER & PD & 0.000674\\ \hline
NLI & SS & 1.05897e-10\\ \hline
NLI & Sum & 0.001194\\ \hline
NLI & WSD & 0.003894\\ \hline
SA & SS & 0.0000256\\ \hline
SS & SRL & 0.00000224\\ \hline
SS & PD & 9.81710e-11\\ \hline
SS & Sum & 0.0165866\\ \hline
SS & WSD & 0.0057237\\ \hline
SS & QA & 2.98083e-7\\ \hline
PD & Sum & 0.000685\\ \hline
PD & WSD & 0.00231873\\
\hline
\end{tabular}
\caption{Pairwise comparison one-way Anova results for DFL-R region}
\label{tab:DFL-R}
\end{table}

\begin{table}[!htb]
\scriptsize
\centering
\begin{tabular}{|c|c|c |}
\hline
T1 & T2& p-value\\ \hline 
CR & NLI & 0.0188070\\ \hline
CR & PD & 0.0099703\\ \hline
NER & PD & 0.0321010\\ \hline
NLI & SS & 6.37802e-7\\ \hline
SA & SS & 0.00642888\\ \hline
SS & SRL & 0.00051148\\ \hline
SS & PD & 2.42829e-7\\ \hline
SS & QA & 0.000162\\ \hline
PD & WSD & 0.0476\\
\hline
\end{tabular}
\caption{Pairwise comparison one-way Anova results for VC-L region}
\label{tab:VC-L}
\end{table}

\begin{table}[!htb]
\scriptsize
\centering
\begin{tabular}{|c|c|c |}
\hline
T1 & T2& p-value\\ \hline 
CR & NLI & 0.00498313\\ \hline
CR & PD & 0.000298933\\ \hline
NER & NLI & 0.024695\\ \hline
NER & PD & 0.0020556\\ \hline
NLI & SS & 4.16645e-8\\ \hline
NLI & Sum & 0.0449825\\ \hline
NLI & WSD & 0.0352242\\ \hline
SA & SS & 0.00120394\\ \hline
SS & SRL & 0.00002939\\ \hline
SS & PD & 7.70669e-10\\ \hline
SS & Sum & 0.0417081\\ \hline
SS & QA & 0.00000742881\\ \hline
PD & Sum & 0.00434934\\ \hline
PD & WSD & 0.0031\\
\hline
\end{tabular}
\caption{Pairwise comparison one-way Anova results for VC-R region}
\label{tab:VC-R}
\end{table}

\subsection{Narratives-Pieman dataset}
The main effect of model was significant for the ROIs with 95\% confidence with these statistics:

\begin{itemize}
\itemsep0em
\item EAC\_L [F(9,810)=3.88, \emph{p}=.00009]
\item EAC\_R [F(9,810)=3.34, \emph{p}=.00055]
\item AAC\_L [F(9,810)=5.37, \emph{p}=.0000007]
\item AAC\_R [F(9,810)=6.955, \emph{p}=.00000] 
\item PMC\_L [F(9,810)=37.21, \emph{p}=.00000]
\item PMC\_R [F(9,810)=31.62, \emph{p}=.00000]
\item TPOJ\_L [F(9,810)=9.166, \emph{p}=.00000]
\item TPOJ\_R [F(9,810)=7.797, \emph{p}=.00000]
\item DFL\_L [F(9,810)=12.445, \emph{p}=.00000]
\item DFL\_R [F(9,810)=12.27, \emph{p}=.00000]
\end{itemize}

\begin{figure*}[t] 
\centering
\includegraphics[width=0.8\linewidth]{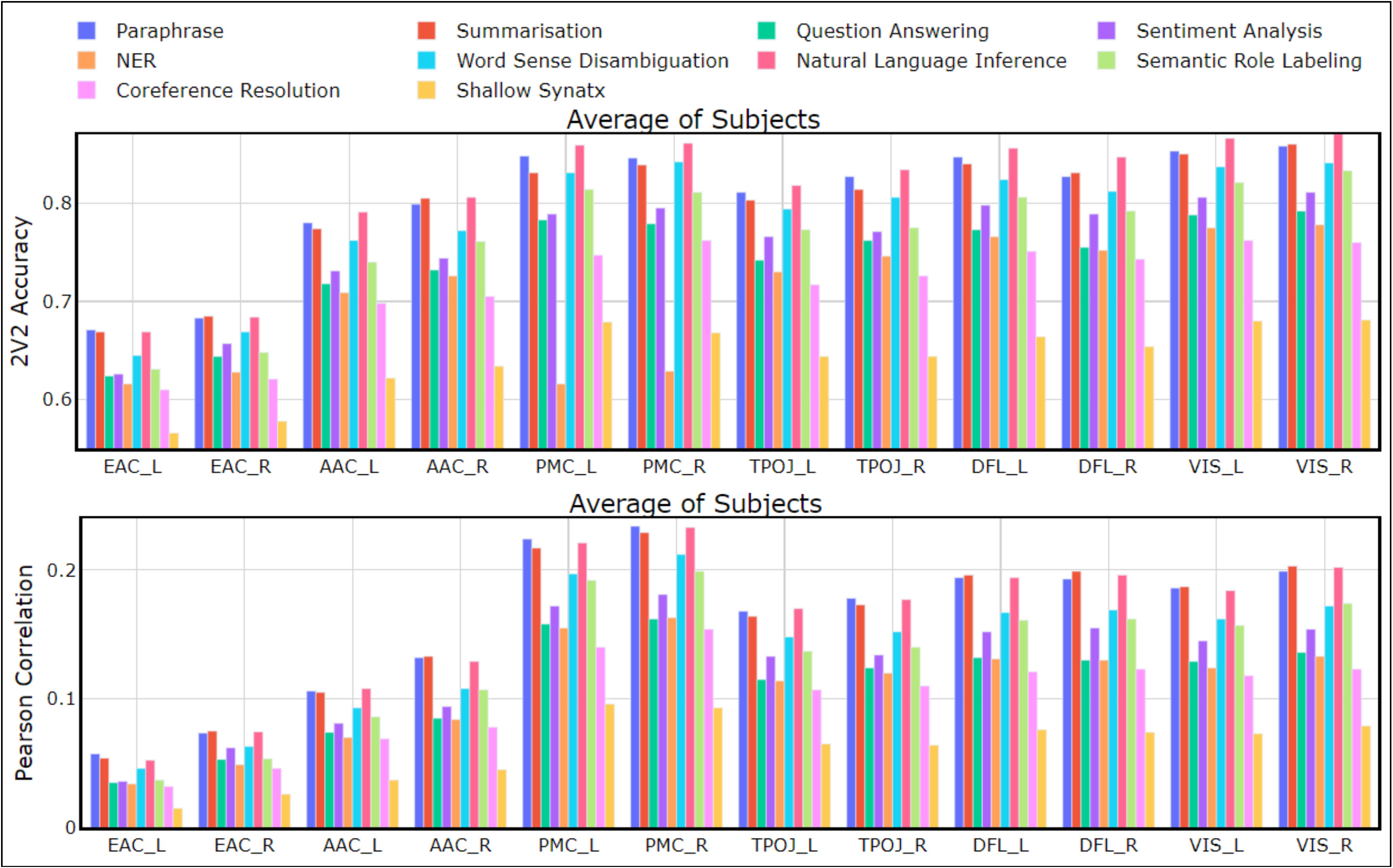}
\caption{Narratives-Lucy Dataset: 2V2 Accuracy (top figure) and Pearson correlation coefficient (bottom figure) between predicted and true responses across different brain regions using a variety of NLP tasks. Results are averaged across all participants. NLI, Paraphrase, and Summarisation perform the best.}
\label{fig:lucy_2v2_pcc_avg}
\end{figure*}

\begin{figure*}[t] 
\centering
\includegraphics[width=0.8\linewidth]{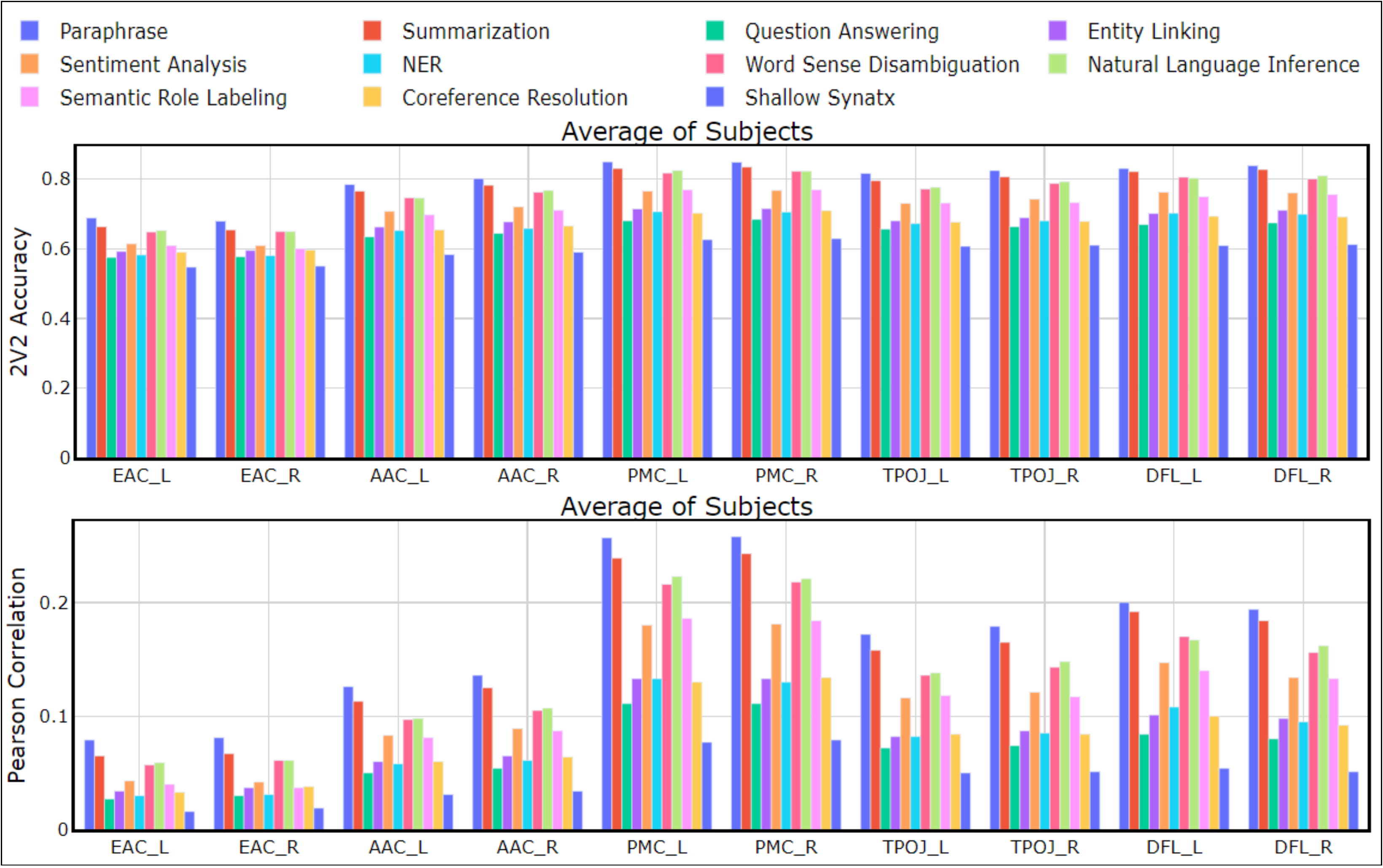}
\caption{Narratives-Slumlord Dataset: 2V2 Accuracy (top figure) and Pearson correlation coefficient (bottom figure) between predicted and true responses across different brain regions using a variety of NLP tasks. Results are averaged across all participants. NLI, Paraphrase, and Summarisation perform the best.}
\label{fig:slumlord_2v2_pcc_avg}
\end{figure*}

\begin{figure}[t] 
\includegraphics[width=\linewidth]{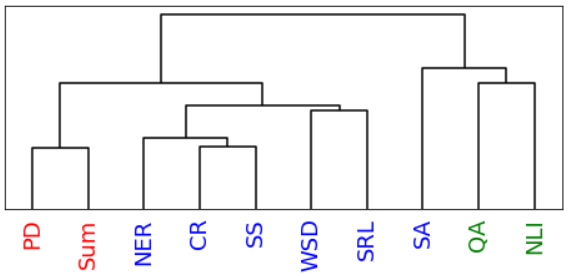}
\includegraphics[width=\linewidth]{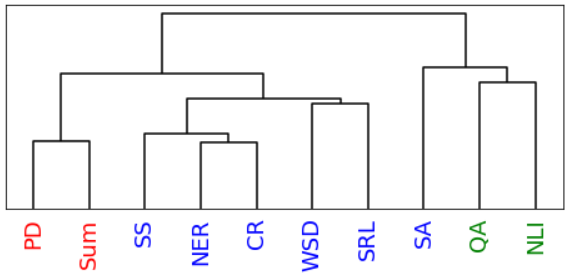}
\includegraphics[width=\linewidth]{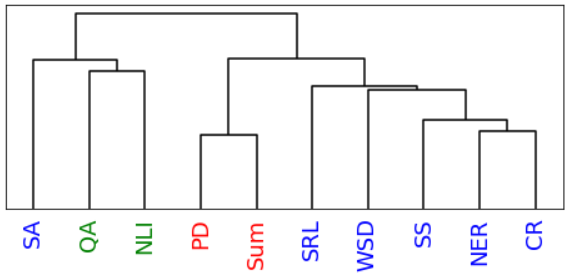}
\caption{Dendrogram constructed using similarity matrix constructed from the task-wise brain response predictions across 10 tasks for subjects 1,  2 and 7 in Pereira Dataset}
\label{fig:Dendrograms_pereira_transfer_subjects}
\end{figure}

\begin{figure}[t] 
\includegraphics[width=\linewidth]{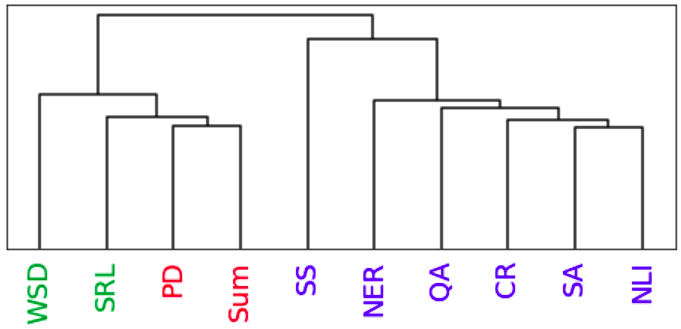}
\includegraphics[width=\linewidth]{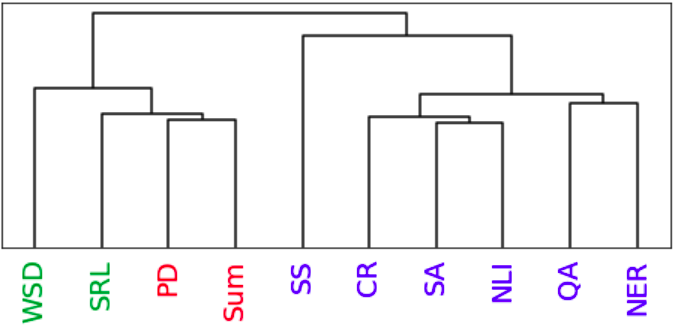}
\includegraphics[width=\linewidth]{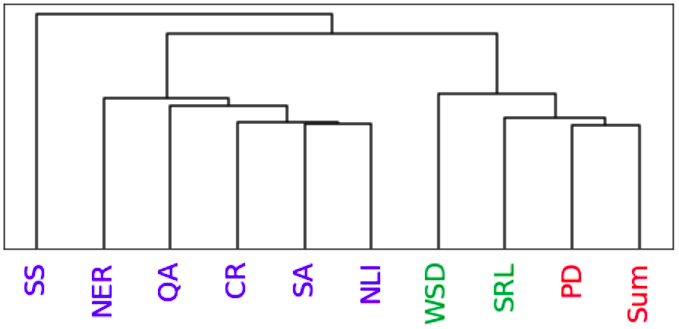}
\caption{Dendrogram constructed using similarity matrix constructed from the task-wise brain response predictions across 10 tasks for subjects 1, 21 and 31 in Narratives Dataset}
\label{fig:Dendrograms_narratives_transfer_subjects}
\end{figure}

\begin{figure}[t]
\centering
\includegraphics[width=0.55\linewidth]{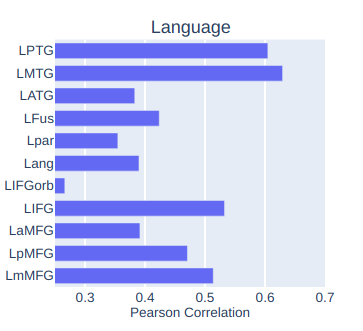}
\caption{Pereira Dataset -- Pearson correlation coefficient between predicted and true responses across different sub ROIs of the Language Network using SRL task. Results are averaged across all participants.}
\label{fig:language_pereira_srl}
\end{figure}

\begin{figure}[t]
\centering
\includegraphics[width=0.55\linewidth]{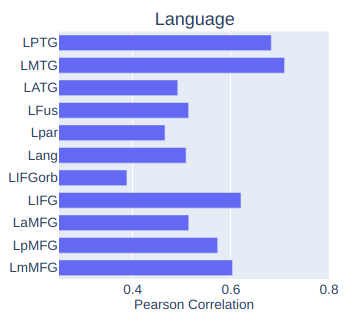}
\caption{Pereira Dataset -- Pearson correlation coefficient between predicted and true responses across different sub ROIs of the Language Network using CR task. Results are averaged across all participants.}
\label{fig:language_pereira}
\end{figure}

\begin{figure}[t]
\centering
\includegraphics[width=0.85\linewidth]{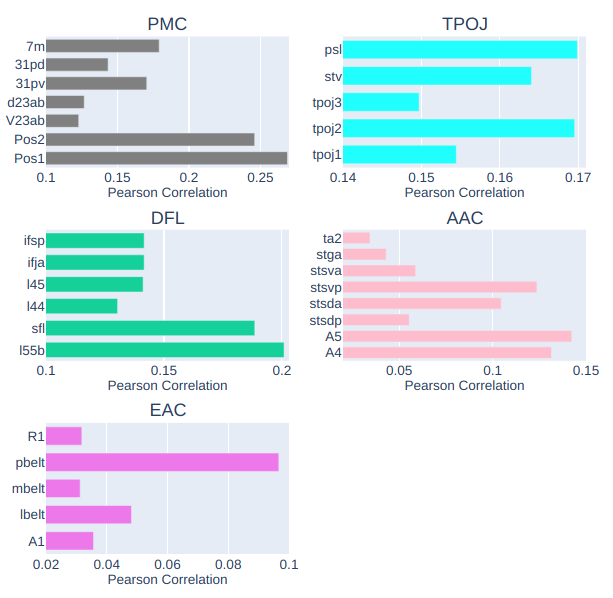}
\caption{Narratives-Pieman -- Pearson correlation coefficient between predicted and true responses across different sub ROIs of 5 brain ROIs using paraphrase task. Results are averaged across all participants.}
\label{fig:subrois_narratives}
\end{figure}

\begin{figure}[t]
\centering
\includegraphics[width=0.85\linewidth]{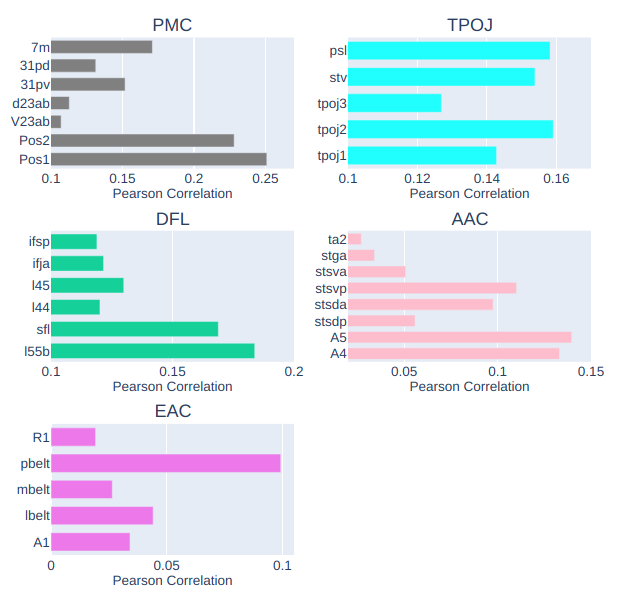}
\caption{Narratives-Pieman -- Pearson correlation coefficient between predicted and true responses across different sub ROIs of 5 brain ROIs using summarization task. Results are averaged across all participants.}
\label{fig:subrois_narratives_summary}
\end{figure}

\begin{figure*}[t] 
\centering
\includegraphics[width=\linewidth]{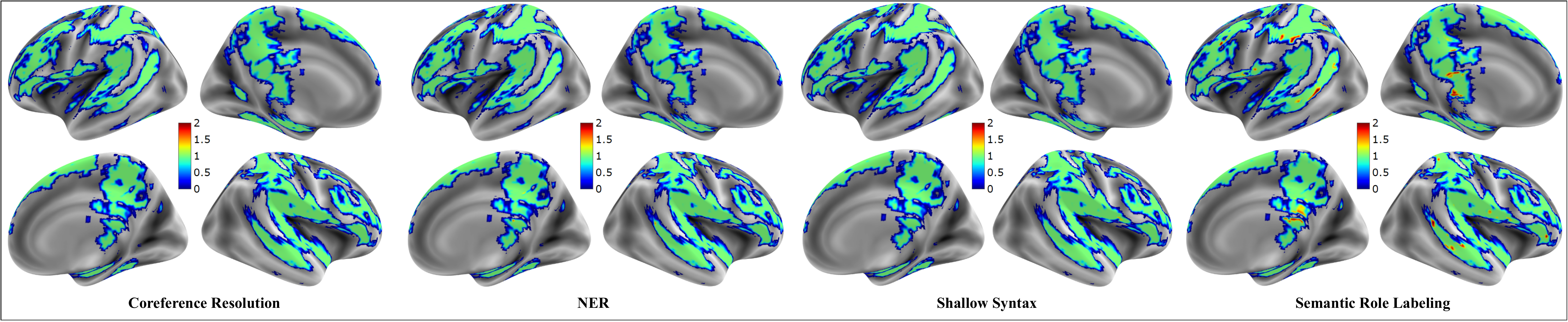}
\includegraphics[width=\linewidth]{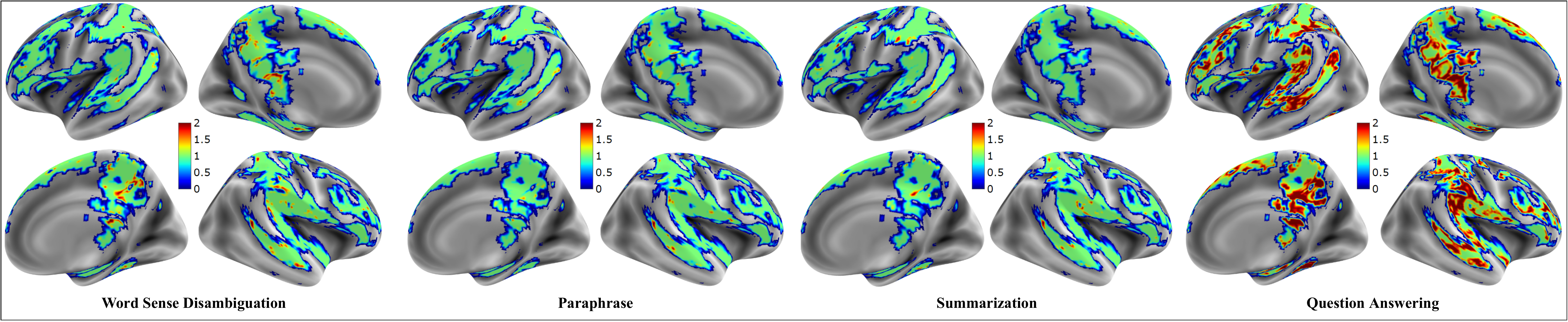}
\includegraphics[width=0.5\linewidth]{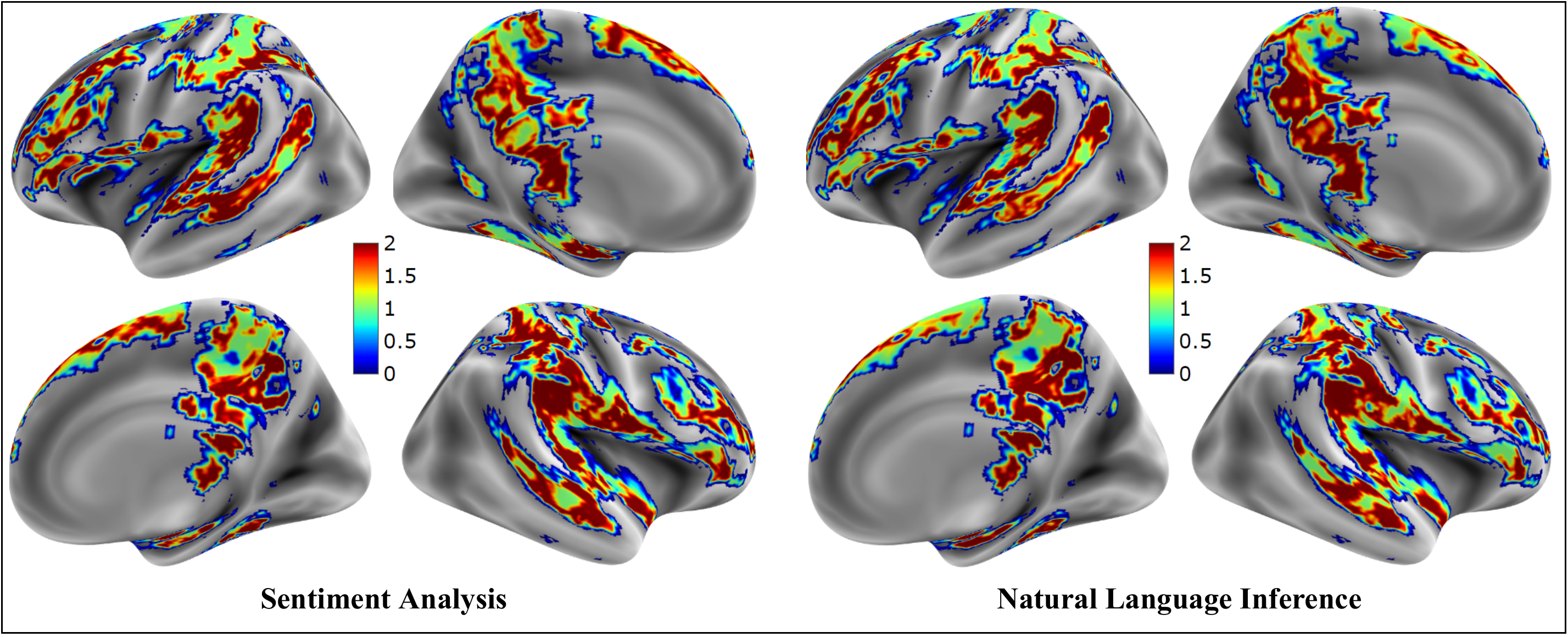}
\caption{Pereira BrainMaps: Mean absolute error (MAE) between predictive voxels and actual voxels using task features from Taskonomy in one sample subject (subject 1). Predictive regions of different tasks are dissimilar across tasks.}
\label{fig:brainmaps_pereira_all}
\end{figure*}

\begin{figure*}[t] 
\centering
\includegraphics[width=\linewidth]{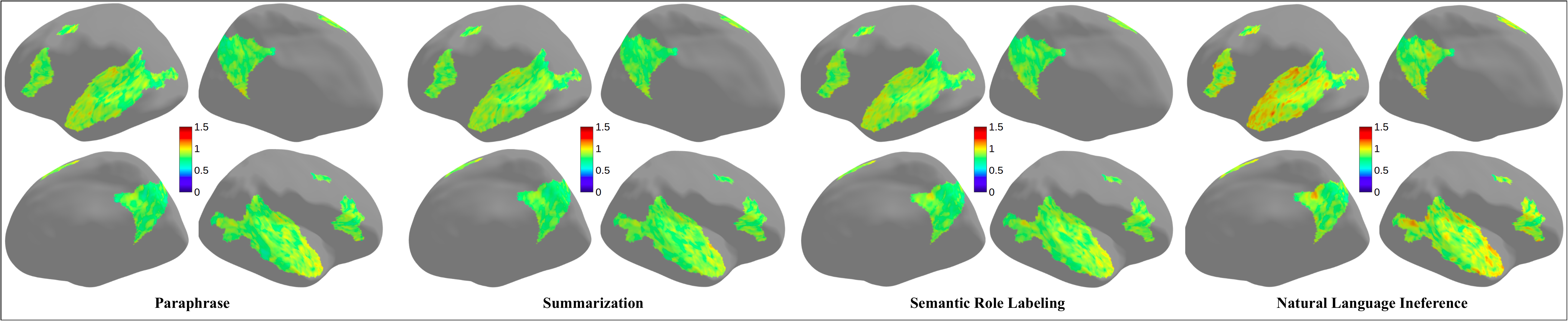}
\includegraphics[width=\linewidth]{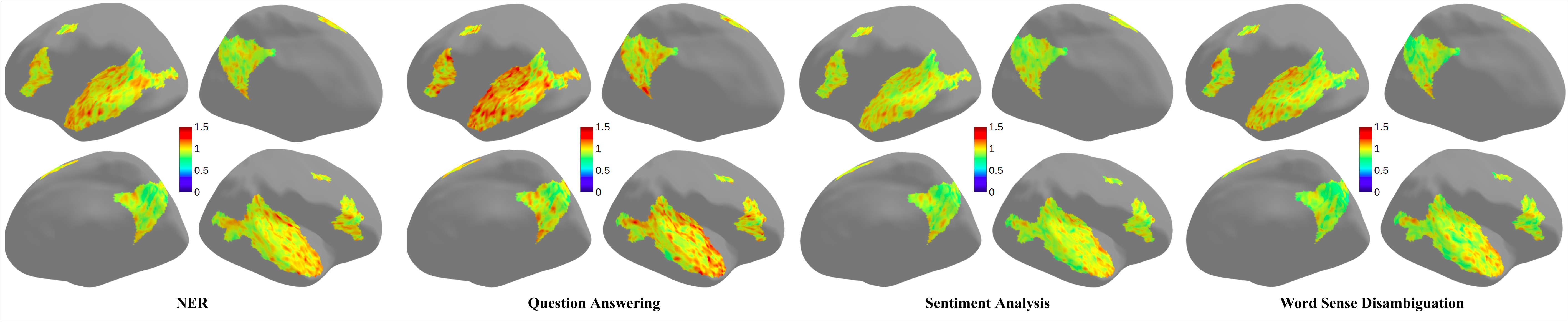}
\includegraphics[width=0.5\linewidth]{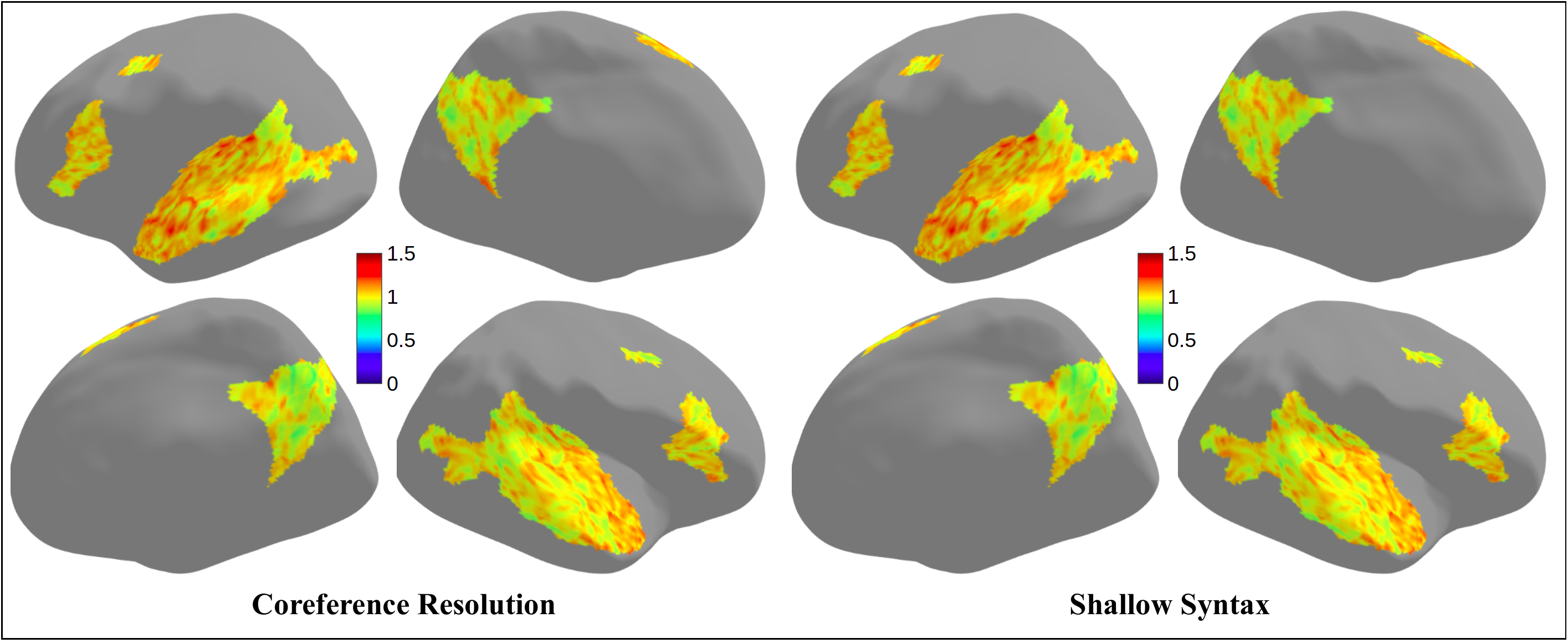}
\caption{Narratives BrainMaps: Mean absolute error (MAE) between predictive voxels and actual voxels using task features from Taskonomy in one sample subject (subject 1) of PieMan dataset. Predictive regions of various tasks are different across tasks.}
\label{fig:brainmaps_narratives_all}
\end{figure*}

\end{document}